%% file: tmi.tex
\documentclass[journal,twoside,web]{ieeecolor}
\usepackage{tmi}
\usepackage{cite}
\usepackage{amsmath,amssymb,amsfonts}
\usepackage{algorithmic}
\usepackage{graphicx}
\usepackage{textcomp}
\usepackage{url}
\usepackage{multirow}

\usepackage[linesnumbered,ruled]{algorithm2e}
\usepackage{color, colortbl}
\usepackage{soul}
\usepackage{spverbatim}
\usepackage{etoolbox}
\usepackage{multirow}
\usepackage{booktabs}
\usepackage{courier}
\usepackage{bm}
\usepackage{gensymb}
\usepackage{diagbox} 
\usepackage{bbm}
\usepackage{array}
\usepackage[caption=false]{subfig}
\usepackage{tabularx,booktabs}

\newcommand{\tablestyle}[2]{\setlength{\tabcolsep}{#1}\renewcommand{\arraystretch}{#2}\centering\small}

\newcommand*{\op}[1]{\operatorname{#1}}

\usepackage[linesnumbered,ruled]{algorithm2e}

\def\BibTeX{{\rm B\kern-.05em{\sc i\kern-.025em b}\kern-.08em
    T\kern-.1667em\lower.7ex\hbox{E}\kern-.125emX}}
\begin{document}
\title{3D TransUNet: Advancing Medical Image Segmentation through Vision Transformers}
\author{Jieneng~Chen,
        Jieru~Mei,
        Xianhang~Li,
        Yongyi~Lu,
        Qihang~Yu,
        Qingyue~Wei,
        Xiangde~Luo,
        Yutong~Xie,
        Ehsan~Adeli,
        Yan~Wang,
        Matthew~Lungren,
        Lei~Xing,
        Le~Lu,
        Alan~Yuille,
        Yuyin~Zhou
\thanks{This work is partially supported by the TPU Research Cloud (TRC) Program, Google Cloud Research Credits Program, the AWS Public Sector Cloud Credit for Research Program, and a 2023 Patrick J. McGovern Foundation Award.}
\thanks{J. Chen, J. Mei, Y. Lu, Q. Yu, A. Yuille are with the Department of Computer Science, Johns Hopkins University, Baltimore, MD 21218, USA (e-mail: jchen293@jh.edu; meijieru@gmail.com; yylu1989@gmail.com; yucornetto@gmail.com; ayuille1@jhu.edu ).}
\thanks{Y. Xie is with the Australian Institute for Machine Learning, University of Adelaide, Australia (e-mail: yutong.xie678@gmail.com).}
\thanks{X. Luo is with the UESTC, Chengdu 610054, China (e-mail: xiangde.luo@std.uestc.edu.cn).}
\thanks{E. Adeli and M. Lungren are with the School of Medicine, Stanford University, Stanford, CA 94305 USA
    (e-mail: eadeli@stanford.edu; mlungren@stanford.edu).}
\thanks{Y. Wang is with the East China Normal University, Shanghai 200062 China (e-mail: wyanny.9@gmail.com).}
\thanks{L. Lu is with the DAMO Academy, Alibaba Group, New York, NY 10014, USA (e-mail: tiger.lelu@gmail.com).}
\thanks{Q. Wei and L. Xing are with the Department of Radiation Oncology, Stanford University, Stanford, CA 94305 USA 
    (e-mail: qywei@stanford.edu; lei@stanford.edu).}
\thanks{X. Li and Y. Zhou are with the Department of Computer Science and Engineering at University of California, Santa Cruz, CA 95064 
    (e-mail: xli421@ucsc.edu; zhouyuyiner@gmail.com).}}
\maketitle

\begin{abstract}

Medical image segmentation plays a crucial role in advancing healthcare systems for disease diagnosis and treatment planning. The u-shaped architecture, popularly known as U-Net, has proven highly successful for various medical image segmentation tasks. However, U-Net's convolution-based operations inherently limit its ability to model long-range dependencies effectively.

To address these limitations, researchers have turned to Transformers, renowned for their global self-attention mechanisms, as alternative architectures. 
One popular network is our previous TransUNet, 
which leverages Transformers' self-attention to complement U-Net's localized information with the global context. In this paper, we extend the 2D TransUNet architecture to a 3D network by building upon the state-of-the-art nnU-Net architecture, and fully exploring Transformers' potential in both the encoder and decoder design. We introduce two key components: 1) A \emph{Transformer encoder} that tokenizes image patches from a convolution neural network (CNN) feature map, enabling the extraction of global contexts, and 2) A \emph{Transformer decoder} that adaptively refines candidate regions by utilizing cross-attention between candidate proposals and U-Net features.
Our investigations reveal that different medical tasks benefit from distinct architectural designs. The Transformer encoder excels in multi-organ segmentation, where the relationship among organs is crucial. On the other hand, the Transformer decoder proves more beneficial for dealing with small and challenging segmented targets such as tumor segmentation.
Extensive experiments showcase the significant potential of integrating a Transformer-based encoder and decoder into the u-shaped medical image segmentation architecture. TransUNet outperforms competitors in various medical applications, including multi-organ segmentation, pancreatic tumor segmentation, and hepatic vessel segmentation. It notably surpasses the top solution in the BrasTS2021 challenge. Code and models are available at~\url{https://github.com/Beckschen/3D-TransUNet}.

\end{abstract}

\section{Introduction}
Convolutional neural networks (CNNs), particularly fully convolutional networks (FCNs)\cite{long2015fully}, have risen to prominence in the domain of medical image segmentation. Among their various iterations, the U-Net model\cite{ronneberger2015u}, characterized by its symmetric encoder-decoder design augmented with skip-connections for improved detail preservation, stands out as the preferred choice for many researchers. Building on this methodology, remarkable progress has been witnessed across diverse medical imaging tasks. These advancements encompass cardiac segmentation in magnetic resonance (MR) imaging~\cite{yu2017automatic}, organ delineation using computed tomography (CT) scans~\cite{zhou2017fixed,li2018h,yu2018recurrent,luo2021semi}, and polyp segmentation in colonoscopy recordings~\cite{zhou2018unet++}.

Despite CNNs' unparalleled representational capabilities, they often falter when modeling long-range relationships due to the inherent locality of convolution operations. This limitation becomes particularly pronounced in cases with large inter-patient variations in texture, shape, and size. 
Recognizing this limitation, the research community has been increasingly drawn to Transformers, models built entirely upon attention mechanisms due to their innate prowess in capturing global contexts~\cite{vaswani2017attention}. 
In the realm of medical image segmentation, our prior work with TransUNet\cite{chen2021transunet} stands as a testament to the potential of transformers.
However, a pivotal observation from our research indicates that simply substituting a CNN encoder with a Transformer can lead to suboptimal outcomes. Transformers process inputs as 1D sequences and prioritize global context modeling, inadvertently producing features of low resolution. Directly upsampling such features fails to reintroduce the lost granularity.
In contrast, a hybrid approach combining CNN and Transformer encoders seems more promising. It effectively harnesses the high-resolution spatial details from CNNs while also benefiting from the global context provided by Transformers.

In this study, we extend the original 2D TransUNet architecture to a 3D configuration, delving deeper into the strategic incorporation of Transformers in both encoding and decoding processes. This leap is rooted in the prowess of the nnU-Net framework, with a vision to surpass its established standards.
Our 3D TransUNet unfolds through two primary mechanisms: Firstly, the Transformer Encoder tokenizes image patches from CNN feature maps,  allowing a seamless fusion of global self-attentive features with high-resolution CNN features skipped from the encoding path, for enabling precise localization. Secondly, the Transformer Decoder redefines conventional per-pixel segmentation as a mask classification, framing prediction candidates as learnable queries. Specifically, these queries are progressively refined by synergizing cross-attention with localized multi-scale CNN features. In addition, we introduce coarse-to-fine attention refinement in the Transformer decoder where for each segmentation class, an initial candidate set is meticulously refined using attention mechanisms focused on the prediction's foreground, ensuring that each iterative refinement sets a new standard for the subsequent, culminating in continuously improved segmentation accuracy.

By integrating Transformers in the encoder and decoder components of the U-Net-like architectures, we show that our designs allow the framework to preserve the advantages of Transformers while enhancing medical image segmentation. Intriguingly, multi-organ segmentation, which leans heavily on global context information—such as the interplay among diverse abdominal organs—tends to gravitate towards the Transformer encoder design. Conversely, tasks like small target segmentation, such as tumor detection, generally benefit more from the Transformer decoder design. 
Our extensive experiments demonstrate the superior performance of our method compared to competing approaches across various medical image segmentation tasks.  In summary, our contributions can be summarized as follows: 
\begin{itemize}
    \item We introduce a Transformer-centric medical image segmentation framework, incorporating self-attention within the sequence-to-sequence prediction context, applicable to both 2D and 3D medical image segmentation tasks. 
    \item We thoroughly investigate the effects of integrating vision transformers into the encoder and the decoder of the u-shaped segmentation architectures, providing insights on tailoring designs to cater to distinct medical image segmentation challenges.
    \item We achieve the state-of-the-art results on various medical image segmentation tasks,  and release our codebase to encourage further exploration in applying Transformers to medical applications.
\end{itemize}

\section{Related Works}
\noindent\textbf{Combining CNNs with self-attention mechanisms.}
Various studies have attempted to integrate self-attention mechanisms into CNNs by modeling global interactions of all pixels based on the feature maps. For instance, Wang \emph{et al.} designed a non-local operator, which can be plugged into multiple intermediate convolution layers~\cite{wang2018non}. Built upon the encoder-decoder u-shaped architecture, Schlemper~\emph{et al.}~\cite{schlemper2019attention} proposed additive attention gate modules which are integrated into the skip-connections.
Different from these approaches, we employ Transformers for embedding global self-attention in our method.

\vspace{1ex}\noindent\textbf{Transformers.}
Transformers were first proposed by \cite{vaswani2017attention} for machine translation and established state-of-the-arts in many NLP tasks.
To make Transformers also applicable for computer vision tasks, several modifications have been made.
For instance, Parmar \emph{et al.}~\cite{parmar2018image} applied the self-attention only in local neighborhoods for each query pixel instead of globally.
Child \emph{et al.}~\cite{child2019generating} proposed Sparse Transformers, which employ scalable approximations to global self-attention.
Recently, Vision Transformer (ViT)~\cite{dosovitskiy2020image} achieved state-of-the-art on ImageNet classification by directly applying Transformers with global self-attention to full-sized images. 
To the best of our knowledge, our TransUNet originally proposed in 2021 is the first Transformer-based medical image segmentation framework, which builds upon the highly successful ViT.
Based on TransUNet, nnformer~\cite{zhou2023nnformer} improves the methodology by interleaving convolution with self-attention. In another development, CoTR~\cite{xie2021cotr} offers a more efficient self-attention in the Transformer encoder.
Another Transformer architecture is the Swin Transformer~\cite{liu2021swin} a hierarchical vision transformer that employs shifted windows to capture local and global information for efficient and scalable visual processing. Subsequent models such as SwinUNet~\cite{cao2022swin} and SwinUETR~\cite{hatamizadeh2021swin} have been developed for medical image segmentation.

\vspace{1ex}\noindent\textbf{Mask classification for segmentation.} DETR~\cite{carion2020end} is the first work that uses Transformer as a decoder with learnable object queries for object detection. In the context of recent advancements in transformers \cite{strudel2021segmenter, wang2021max, cheng2021per, cheng2022masked, yu2022k, yu2022cmt}, a novel variation known as mask Transformers has emerged. This variant introduces segmentation predictions by employing a collection of query embeddings to represent object and its associated mask. Wang \emph{et al.}~\cite{wang2021max} first develops mask Transformer with memory embedding, and Cheng \emph{et al.}~\cite{cheng2021per} further formulate the query update in a manner of DETR~\cite{carion2020end}.  At the core of mask transformers lies the decoder, which is responsible for processing object queries as input and progressively transforming them into mask embedding vectors~\cite{cheng2021per, cheng2022masked, yu2022k, yu2022cmt}. This process enables the model to effectively handle segmentation tasks and produce accurate results.

\section{Method}
Given a 3D medical image (\emph{e.g.}, CT/MR scan) $\bm{\mathrm{x}}\in \mathbb{R}^{D \times H \times W \times C}$ with the spatial resolution of $D \times H \times W$ and $C$ number of channels. Our goal is to predict the corresponding pixel-wise labelmap with size $D \times H \times W$. The most common way is to directly train a CNN (\emph{e.g.}, U-Net) to first encode images into high-level feature representations, which are then decoded back to the full spatial resolution.
Our approach diverges from conventional methods by thoroughly exploring the attention mechanisms utilized in both the encoder and decoder phases of standard U-shaped segmentation architectures, employing Transformers. 
In Section~\ref{sec:transformer_encoder}, we delve into the direct application of Transformers for encoding feature representations from segmented image patches. Following this, in Section~\ref{sec:transformer_decoder}, we elaborate on the implementation of the query-based Transformer, which serves as our decoder. The detailed architecture of TransUNet is then presented in Section~\ref{sec:transunet}.

\subsection{Transformer as Encoder}
\label{sec:transformer_encoder}
\noindent\textbf{Image sequentialization.} 
Following~\cite{dosovitskiy2020image}, we first perform tokenization by reshaping the input $\bm{\mathrm{x}}$ into a sequence of flattened 3D patches \{$\bm{\mathrm{x}}^p_i \in \mathbb{R}^{P^3 \cdot C}|i=1,..,N\}$, where each patch is of size $P \times P \times P$ and $N=\frac{DHW}{P^3}$ is the number of image patches (\emph{i.e.}, the input sequence length). 

\vspace{1ex}\noindent\textbf{Patch embedding.} We map the vectorized patches $\bm{\mathrm{x}}^p$ into a latent $d_{enc}$-dimensional embedding space using a trainable linear projection.
To encode the patch spatial information, we learn specific position embeddings which are added to the patch embeddings to retain positional information as follows:
\begin{align}
    \bm{\mathrm{z}}_0 &= [\bm{\mathrm{x}}^p_1 \bm{\mathrm{E}}; \, \bm{\mathrm{x}}^p_2 \bm{\mathrm{E}}; \cdots; \, \bm{\mathrm{x}}^{p}_N \bm{\mathrm{E}} ] + \bm{\mathrm{E}}^{pos}, \label{eq:embedding} 
\end{align}

\noindent where $\bm{\mathrm{E}} \in \mathbb{R}^{(P^3 \cdot C) \times d_{enc}}$ is the patch embedding projection, and $\bm{\mathrm{E}}^{pos}  \in \mathbb{R}^{N \times d_{enc}}$ denotes the position embedding.

Each Transformer layer consists of Multihead Self-Attention (MSA) and Multi-Layer Perceptron (MLP) blocks (Eq.~\eqref{eq:msa_apply}\eqref{eq:mlp_apply}). Therefore the output of the $\ell$-th layer can be written as follows:
\begin{align}
    \bm{\mathrm{z}}^\prime_\ell &= \op{MSA}(\op{LN}(\bm{\mathrm{z}}_{\ell-1})) + \bm{\mathrm{z}}_{\ell-1}, &&  \label{eq:msa_apply} \\
    \bm{\mathrm{z}}_\ell &= \op{MLP}(\op{LN}(\bm{\mathrm{z}}^\prime_{\ell})) + \bm{\mathrm{z}}^\prime_{\ell},   \label{eq:mlp_apply} 
\end{align}

where $\op{LN}(\cdot)$ denotes the layer normalization operator and $\bm{\mathrm{z}}_\ell$ is the encoded image representation.

\subsection{Transformer as Decoder}
\label{sec:transformer_decoder}

\subsubsection{Coarse candidate estimation} Traditional approaches, such as U-Net, predominantly view medical image segmentation as a per-pixel classification task. In this paradigm, each pixel is classified into one of the possible \( K \) categories, typically achieved by training a segmentation model with the per-pixel cross-entropy (or negative log-likelihood) loss.

Instead of considering individual pixels, our approach in this paper treats medical image segmentation as a mask classification problem. 
We introduce the concept of an ``organ query'', a 
$d_{dec}$-dimensional feature vector that represents each organ in the image. With a predefined set of $N$ organ queries, our goal for an image comprising $K$ segmentation classes is to segregate the image into 
$N$ distinct candidate regions. Subsequently, we aim to assign the corresponding organ label to each region.
Importantly, it is worth noting that the value of 
$N$ does not have to align with the number of classes, as demonstrated in prior studies~\cite{strudel2021segmenter}. In fact, we intentionally set $N$ to be significantly larger than $K$, to minimize the risk of false negatives.
Assume the dimension of the object queries is $d_{dec}$, the coarse predicted segmentation map can be computed by the dot product between the initial organ queries $\mathbf{P}^0 \in \mathbb{R}^{N\times d_{dec}}$ and the embedding of the U-Net last block feature $\mathbf{F} \in \mathbb{R}^{D \times H \times W\times d_{dec}}$: 
\begin{align}
\label{eq:coarse_prediction}
& \mathbf{Z}^0 = g(\mathbf{P}^0 \times \mathbf{F}^\top),
\end{align}
where $g(\cdot)$ is sigmoid activation followed by a hard thresholding operation with a threshold set at 0.5.
As shown in Eq.~\eqref{eq:coarse_prediction}, $\mathbf{P}$ can be seen as the 1$\times$1$\times$1 convolutional kernel in the standard U-Net segmentation head.

\subsubsection{Transformer decoder}
Figure~\ref{fig:framework} illustrates our customized 3D Transformer decoder designed specifically to refine organ queries, thereby enhancing the coarse prediction $\mathbf{Z}^0$.
Similar to the structure seen in the Transformer encoder (detailed in Section~\ref{sec:transformer_encoder}), the self-attention mechanism ($i.e.,$ the MSA block) in each layer will enable the Transformer decoder 
to comprehensively engage with image features and capture organ query interrelations. Recognizing the rich localization in intermediate CNN features, which complements the Transformer's global image context, we refine organ queries in each decoder layer by integrating cross-attention with localized multi-scale CNN features.

Our strategy involves concurrent training of a CNN decoder and the Transformer decoder. In the \(t\)-th layer, the refined organ queries are denoted as \(\mathbf{P}^t \in \mathbb{R}^{N\times d_{dec}}\). Simultaneously, an intermediate U-Net feature is mapped to a \(d_{dec}\)-dimensional feature space denoted as \(\mathcal{F}\) to facilitate cross-attention computation. Notably, when the count of upsampling blocks aligns with the Transformer decoder layers, multi-scale CNN features can be projected into the feature space \(\mathcal{F} \in \mathbb{R}^{(D_t H_t W_t) \times d_{dec}}\), where \(D_t\), \(H_t\), and \(W_t\) specify the spatial dimensions of the feature map at the \(t\)-th upsampling block.
Moving to the \(t+1\)-th layer, organ queries are updated using cross-attention as follows:
\begin{align}
\label{eq:coarse_candidate_update}
    & \mathbf{P}^{t+1} = \mathbf{P}^{t} + \text{Softmax}((\mathbf{P}^{t}\mathbf{w}_{q})(\mathcal{F}\mathbf{w}_{k})^\top)\times \mathcal{F}\mathbf{w}_{v},
\end{align}
where the \(t\)-th query features undergo linear projection to form queries for the next layer using the weight matrix \(\mathbf{w}_{q} \in \mathbb{R}^{d_{dec} \times d_q}\). The U-Net feature, \(\mathcal{F}\), is similarly transformed into keys and values using parametric weight matrices \(\mathbf{w}_{k} \in \mathbb{R}^{d_{dec} \times d_k}\) and \(\mathbf{w}_{v} \in \mathbb{R}^{d_{dec} \times d_v}\). Note a residual path is used for updating $\mathbf{P}$ following previous studies~\cite{cheng2022masked}. 
Next, we will introduce how to incorporate a coarse-to-fine attention refinement to further enhance the accuracy of segmentation results.

\subsubsection{Coarse-to-fine attention refinement}
The value of coarse-to-fine refinement in medical image segmentation, particularly for small target segmentation, is well-established~\cite{zhou2017fixed, zhu20183d, xie2019recurrent}. This technique employs a coarse mask from an initial stage to guide subsequent refinements. Here, to integrate a seamless coarse-to-fine refinement process within the Transformer decoder, we have incorporated a mask attention module~\cite{cheng2022masked}. This enhancement aims to ground the cross-attention within the foreground region based on the former coarse prediction for each category, to reduce the background noise and better focus on the region of interest. This improved attention map iteratively aids subsequent, finer segmentation stages.

Concretely, we start by setting the organ queries and the coarse-level mask prediction as $\mathbf{P}^0$ and $\mathbf{Z}^0$ (based on Eq.~\eqref{eq:coarse_prediction}) respectively, and then begin the iterative refinement process. 
At the $t$-th iteration, 
using the current organ query features  $\mathbf{P}^t$ and coarse prediction $\mathbf{Z}^t$, we compute the masked cross-attention, which refines  $\mathbf{P}^{t+1}$
for the subsequent iteration. This computation incorporates the existing coarse prediction $\mathbf{Z}^t$ into the affinity matrixn, as detailed in Eq.\eqref{eq:coarse_candidate_update}:
\begin{align}
\label{eq:query_update_c2f_attention}
  \mathbf{P}^{t+1} = \mathbf{P}^{t} + \text{Softmax}((\mathbf{P}^{t}\mathbf{w}_{q})(\mathcal{F}\mathbf{w}_{k})^\top + h(\mathbf{Z}^{t}))\times \mathcal{F}\mathbf{w}_{v},
\end{align}
where 
\begin{equation}
h\left(\mathbf{Z}^t(i, j, s)\right)=\left\{\begin{array}{ll}
0 & \text { if } \mathbf{Z}^t(i, j, s)=1 \\
-\infty & \text { otherwise }
\end{array}\right.
\end{equation}
where $i, j, s$ are the coordinate indices. This formula restricts the cross-attention mechanism to focus solely on the foreground, nullifying it for all other regions.
By iteratively updating both the organ queries and the corresponding mask predictions, our Transformer decoder systematically refines the segmentation results across multiple iterations. A detailed description of this iterative process is outlined in Algorithm~\ref{Alg:coarse_to_fine}. The refinement cycle persists until the iteration count $t$ reaches the maximum threshold $T$, which is equivalent to the number of layers in the Transformer decoder.

\begin{algorithm}[t!]
\SetKwInOut{Input}{Input}
\SetKwInOut{Output}{Output}
\SetKwInOut{Return}{Return}
\Input{
    Parametric weight matrices $\mathbf{w}_{q},\mathbf{w}_{k},\mathbf{w}_{v}$;\\
    ~Organ embedding $\mathbf{P}$, U-Net last feature $\mathbf{F}$; \\
    ~The U-Net $t$-th layer feature $\mathcal{F}$;\\
    ~Max number of iterations $T$;\\
}
\Output{
    Fine segmentation map $\mathbf{Z}^T$, predicted\\ ~class label $\hat{\mathbf{y}}$;
}
${t}\leftarrow{0}$; \\
$\mathbf{P}^{0}\leftarrow{\mathbf{P}}$;\\
$\mathbf{Z}^{0}\leftarrow{g(\mathbf{P}^{0}\times {\mathbf{F}^\top}})$;\\
\Repeat{${t}={T}$\ }{
    Update $\mathbf{P}^{t+1}$ according to Eq.~\eqref{eq:query_update_c2f_attention};\\
    Update $\mathbf{Z}^{t+1}\leftarrow{g(\mathbf{P}^{t+1}\times {\mathbf{F}^\top}})$; \\
    ${t}\leftarrow{t+1}$;\\
}
Compute the class label $\hat{\mathbf{y}}$ by Eq.~\eqref{eq:class_digits} and Eq.~\eqref{eq:class_label};\\
\Return{
    $\mathbf{Z}^T$, $\hat{\mathbf{y}}$.
}
\caption{
    Iterative coarse-to-fine refinement
}
\label{Alg:coarse_to_fine}
\end{algorithm}

\vspace{1ex}\noindent\textbf{Fine segmentation decoding.} 
After the final iteration, the updated organ queries $\mathbf{P}^T$ can be decoded back to the finalized refined binarized segmentation map $\mathbf{Z}^T$ by the dot product with U-Net's last block feature $\mathbf{F}$, following Eq.~\eqref{eq:coarse_prediction}. 
To associate each binarized mask with one semantic class, we further use a linear layer with weight matrices $\mathbf{w}_{fc} \in \mathbb{R}^{d \times K}$ that projects the refined organ embedding $\mathbf{P}^T$ to the output class logits $\mathbf{O} \in  \mathbb{R}^{N \times K}$. 
Formally, we have:
\begin{align}
\label{eq:class_digits}
    & \mathbf{O} = \mathbf{P}^T \mathbf{w}_{fc} \\
\label{eq:class_label}
    & \hat{\mathbf{y}} = \mathrm{arg\,max}_{k=0,1,...,K-1} ~\mathbf{O} 
\end{align}
where k is the label index.
The final class labels associated with the refined predicted masks $\mathbf{Z}^T$ is $ \hat{\mathbf{y}} \in \mathbb{R}^{N}$.


\input{table_figures/fig_decoder}

\subsection{TransUNet}
\label{sec:transunet}
Our 3D TransUNet is built
upon the state-of-the-art nnU-Net architecture, with the aim of surpassing its established standards.
We illustrate the overall framework of the proposed TransUNet in Figure~\ref{fig:framework}. 
We instantiate our 3D TransUNet with three distinct configurations:
\subsubsection{Encoder-only} 
\textbf{A CNN-Transformer hybrid encoder} is employed where CNN is first used as a feature extractor to generate a feature map for the input. Patch embedding is applied to feature patches instead of from raw images. For the decoding phase, we use a standard U-Net decoder.
We choose this design since 1) it allows us to leverage the intermediate high-resolution CNN feature maps in the decoding path; and 2) we find that the hybrid CNN-Transformer encoder performs better than simply using a pure Transformer as the encoder. The Encoder-only model will be trained using a hybrid segmentation loss consisting of pixel-wise cross entropy loss and dice loss.

\subsubsection{Decoder-only}
In this configuration, we use a conventional CNN encoder for the encoding phase.  As for the decoding phase, we use a \textbf{CNN-Transformer hybrid decoder} in the segmentation model. 
Initially, the organ queries $\mathbf{P}$ are set to zero. Before being processed by the Transformer decoder, they are augmented with learnable positional embeddings following Eq.~\eqref{eq:embedding}. 
Then as aforementioned in Section~\ref{sec:transformer_decoder}, $\mathbf{P}$ will be gradually refined conditioned on the U-Net features and be decoded back into the full-resolution segmentation map.
We train the network with the Hungarian matching loss
following previous works~\cite{carion2020end, wang2021max} to update the organ queries throughout the decoding layers.
This loss aims to match pairs between predictions and ground-truth segments.
It combines pixel-wise classification loss and binary mask loss for each segmented prediction:
\begin{align}
\label{eq:matching_criterion}
\mathcal{L} = \lambda_{0}(\mathcal{L}_{ce} + \mathcal{L}_{dice)} + \lambda_{1}\mathcal{L}_{cls},
\end{align}
where the pixel-wise classification loss $\mathcal{L}_{ce}$ and $\mathcal{L}_{dice}$   denote binary cross-entropy loss and dice loss, respectively~\cite{milletari2016v}.  The classification loss $\mathcal{L}_{cls}$ is instantiated by the cross-entropy loss for each candidate region. $\lambda_{0}$ and $\lambda_{1}$ are the hyper-parameters for balancing the per-pixel segmentation loss and the mask classification loss.

We also employ deep supervision, applying the training loss to the output at each stage of the TransUNet decoder.

\subsubsection{Encoder + Decoder} Here, we integrate both the Transformer encoder and the Transformer decoder into the 3D nnU-Net model. And then similar to the decoder-only model, here we also use the Hungarian matching loss to train the whole network.

Lastly, we would like to note that our method, though built upon the 3D nnU-Net, can be easily modified to fit 2D tasks by simply switching the backbone model and reducing all operations back to 2D.

\section{Experiments and Discussion}
\subsection{Dataset and Evaluation}
\noindent\textbf{Synapse multi-organ segmentation dataset\footnote{\url{https://www.synapse.org/\#!Synapse:syn3193805/wiki/217789}}}.
We use the 30 abdominal CT scans in the MICCAI 2015 Multi-Atlas Abdomen Labeling Challenge, with $3779$ axial contrast-enhanced abdominal clinical CT images in total. 

Each CT volume consists of $85\sim198$ slices of $512\times512$ pixels, with a voxel spatial resolution of $([0.54\sim0.54]\times[0.98\sim0.98]\times[2.5\sim5.0])\textup{mm}^3$. Following~\cite{fu2020domain}, we report the average DSC and average Hausdorff Distance (HD) on 8 abdominal organs (aorta, gallbladder, spleen, left kidney, right kidney, liver, pancreas, spleen, stomach with a random split of 18 training cases (2212 axial slices) and 12 cases for validation, following the split setting in~\cite{chen2021transunet}.



\vspace{1ex}\noindent\textbf{BraTS2021 brain tumor segmentation challenge\footnote{\url{http://braintumorsegmentation.org/}}.} BrasTS2021 Challenge is the most recent and largest dataset for brain tumor segmentation. 1251 multi-parametric MRI scans were provided with segmentation labels to the participants. 4 contrasts are available for the MRI scans: Native T1-weighted image, post-contrast T1-weighted (T1Gd), T2-weighted, and T2 Fluid Attenuated Inversion Recovery (T2-FLAIR). Annotation were manually performed by one to four raters, with final approval from experienced neuro-radiologists. The labels include regions of GD-enhancing tumor (ET), the peritumoral edematous/invaded tissue (ED), and the necrotic tumor core (NCR). All MRI scans were pre-processed by co-registration to the same anatomical template, interpolation to isotropic 1mm$^3$ resolution and skull-stripping. The image sizes of all MRI scans and associated labels are 240$\times$240$\times$155. In our experiments, we apply 5-fold cross-validation with the same data split used by the No.1 solution~\cite{luu2021extending} in BraTS2021 challenge.

\vspace{1ex}\noindent\textbf{Medical Segmentation Decathlon (MSD) HepaticVessel\footnote{\url{http://medicaldecathlon.com/}}.}  
The MSD HepaticVessel, a task of Medical Segmentation Decathlon~\cite{antonelli2021medical}, consists of 443 portal venous phase CT scans obtained from patients with a variety of primary and metastatic liver tumors. The corresponding target ROIs were the vessels and tumors within the liver. This data set was selected due to the tubular and connected nature of hepatic vessels neighboring heterogeneous tumors.  We apply 5-fold cross-validation to evaluate the methods on this dataset.

\vspace{1ex}\noindent\textbf{Large scale pancreatic mass dataset.}  
Our dataset of venous phase 2930 CT scans, is collected from a high-volume US hospital. To the best of our knowledge, it is one of the largest scale pancreatic tumor CT dataset. Pancreatic ductal adenocarcinoma (PDAC) is of the highest priority among all pancreatic abnormalities with a 5-year survival rate of approximately 10\% and is the most common type (about 90\% of all pancreatic cancers). The labels include Pancreas, PDAC and Cyst. The dataset is randomly split into a training of 2123 CT scans and a testing dataset of 807 CT scans. The model validation is conducted on a subset of training set. The training set includes 1017 PDACs, 462 Cyst, and 644 normal pancreases. The testing set includes 506 PDACs, 271 Cyst, 300 normal pancreases. The evaluation metrics include the dice score, the sensitivity and the specificity following the criterion in~\cite{zhu2019multi}.

\vspace{1ex}\noindent\textbf{BraTS2023 brain metastases dataset.} The BraTS-MET dataset~\cite{moawad2023brain4} plays a crucial role in developing advanced algorithms for detecting and segmenting brain metastases, with the goal of ensuring seamless integration into clinical practice. This dataset comprises a variety of untreated brain metastasis mpMRI scans collected from multiple institutions~\cite{oermann2023longitudinal,rudie2023university,grovik2020deep} following standard clinical protocols. It is important to mention that our work exclusively utilizes the BraTS2023-MET version of the dataset. Currently, it consists of 238 cases, and we employ a default 5-fold cross-validation method in our analysis.  

\subsection{Implementation Details}
\label{sec:imple}
\vspace{1ex}\noindent\textbf{Training.} 
In all our 3D experiments, we adhere to nnU-Net's prescribed data augmentation procedures to enhance the diversity of our training dataset. To facilitate effective training, we employ a batch size of 2 using 1 Nvidia RTX 8000 GPU. A comprehensive breakdown of our implementation details can be found in Table~\ref{tab:implement}, encompassing critical aspects such as architectural hyperparameters, training configurations, and data augmentation techniques customized for various datasets.
We experiment with both 1-layer and 12-layer ViT for implementing the Transformer encoder. Specifically, the 12-layer ViT model is pretrained on the ImageNet21k dataset~\cite{russakovsky2015imagenet}, with additional LayerScale~\cite{touvron2021going}. 
The latent dimensions $d_{enc}$ and $d_{dec}$ are set as 768 and 192 respectively.
For computing the Hungarian matching loss, $\lambda_{0}$ and $\lambda_{1}$ are set as 0.7 and 0.3.

In line with nnU-Net's pioneering adaptive architecture design, our TransUNet exhibits adaptability tailored to the characteristics of the data it processes. Similar to nnU-Net, our backbone architecture is inherently self-adaptive for different medical datasets, particularly in determining crucial factors such as the number of down-sampling layers and the allocation of channels at each stage. The details can be found in Table~\ref{tab:implement}.
It is noteworthy that our TransUNet maintains compatibility with 2D image data, requiring minimal architectural modifications to accommodate various imaging scenarios.

\vspace{1ex}\noindent\textbf{Testing.}
Given a CT/MR scan, we do inference in a sliding-window manner. By leveraging the aggregation of all patches, we assign a probability vector to a voxel in position $(i, j, s)$: $\sum_{n=1}^N (\mathbf{Z}^T_{n,ijs}) \in \mathbb{R}^{K}$, followed by an argmax to obtain hard prediction.

\input{table_figures/tab_implement}

\begin{table*}
\centering
\footnotesize
\renewcommand\tabcolsep{1.0pt}
\caption{Comparison on the Synapse multi-organ CT dataset (average dice score \%, and dice score \% for each organ). We use pretrained model from~\cite{dosovitskiy2020image} for the encoder with 12 Transformer layers.
}
\resizebox{0.8\linewidth}{!}
{
\begin{tabular}{cc|c|ccccccccc}
\toprule[0.1em]
\multicolumn{2}{c|}{encoder}  &\multirow{2}{*}{decoder}  & \multirow{2}{*}{Aorta} & \multirow{2}{*}{Gallbladder} & \multirow{2}{*}{Kidney (L)} & \multirow{2}{*}{Kidney (R)} & \multirow{2}{*}{Liver} & \multirow{2}{*}{Pancreas} & \multirow{2}{*}{Spleen} & \multirow{2}{*}{Stomach} & \multirow{2}{*}{Avg. Dice}\\
\cline{1-2}
1-layer &12-layer & & & & & & & & & & \\

\midrule[0.08em]
& &  & 93.04        & 78.82      & 84.68          & 88.46         & 97.13        & 81.50       & 91.68        & 83.34  & 87.33  \\
\checkmark &  &  & 93.07 & 79.56 & 86.16 & 87.68 & 97.22 & 81.71 & 92.56 & 83.23 & 87.65 \\
 & \checkmark &  & 92.97 & 81.15 & 85.76 & 87.47 & 97.03 & 81.76 & 93.39 & 85.31 &\textbf{88.11}\\
  &  & \checkmark  & 92.88 & 82.06 & 86.04 & 87.70 & 97.10 & 82.08 & 91.14 & 82.03 & 87.63 \\
 & \checkmark &\checkmark  & 92.67 & 81.66 & 85.29 & 87.76 & 97.34 & 82.69 & 91.90 & 85.59 & \textbf{88.11} \\
\bottomrule[0.1em]
\end{tabular}
}
\label{tab:synapse-ablation}
\end{table*}

\begin{table}[]
\centering
\caption{Ablation of Transformer architecture on MSD vessel dataset with dice score metrics (\%). Experiments are conducted in five-fold cross-validation.}
\resizebox{1.0\linewidth}{!}
{
\begin{tabular}{cc|c|ccc}
\toprule[0.1em]
\multicolumn{2}{c|}{encoder}  & \multirow{2}{*}{decoder}     & \multirow{2}{*}{Vessel} &\multirow{2}{*}{Tumor} &\multirow{2}{*}{Avg. Dice} \\ 
\cline{1-2}
1-layer &12-layer & & & & \\
\midrule[0.08em]
& & &63.71 &68.36 &66.04 \\
\checkmark &  & & 63.47 & 69.12 & 66.30 \\
 & \checkmark & & 63.67 & 69.02 & 66.35  \\
 &  & \checkmark & 64.41 & 70.94 & \textbf{67.67} \\ 
 \checkmark &  &\checkmark & 64.58  & 69.89 & 67.24   \\
\bottomrule[0.1em]
\end{tabular}
}
\label{tab:vessel-ablation}
\end{table}

\input{table_figures/dsc_msdvessel}



\input{table_figures/dsc_brats23_mets}


\input{table_figures/dsc_felix}

\subsection{Analytical Study of Three Configurations}
\label{sec:analytical}

We performed a variety of ablation studies to thoroughly evaluate the proposed three configurations under the TransUNet framework, $i.e.,$ Encoder-only, Decoder-only and Encoder+Decoder.

To assess how effective Transformer encoders are against CNN encoders, and likewise for decoders, we conducted comprehensive experiments of our 3D TransUNet model. This evaluation encompassed both Encoder-only and Decoder-only configurations and compared them to the baseline nnU-Net across multi-organ segmentation and hepatic vessel tumor segmentation tasks. The results of these comparisons on 3D TransUNet are summarized in Tables~\ref{tab:synapse-ablation} and ~\ref{tab:vessel-ablation}.

For multi-organ segmentation, while the decoder-only design demonstrates a modest performance enhancement (87.63\% compared to 87.33\%), the encoder-only configuration, especially when employing the 12-layer ViT encoder initialized with pre-trained weights from ImageNet, achieves a significant 0.8\% improvement in Dice score, reaching 88.11\%. This advantage becomes even more pronounced when adopting a 2D U-Net backbone, with performance gains of 2.80\% and 3.02\% at resolutions of 224$\times$224 and 512$\times$512, respectively, as outlined in Table~\ref{tab:synapse_comp_sota}.

As for the vessel tumor segmentation, the encoder-only design's performance improvement, while present, remains relatively subtle. Both the 1-layer and 12-layer ViT encoders yield comparable results (66.30\% and 66.35\%), slightly outperforming the baseline nnU-Net's score of 66.04\%. In contrast, the decoder-only configuration exhibits a substantial increment, recording a gain of 1.63\% (67.67\% versus 66.04\%).  

To summarize, our results show the encoder-only design thrives in multi-organ segmentation, while the decoder-only configuration is more adept in vessel tumor segmentation. This distinction aligns with the inherent strengths of the Transformer encoder, which captures global context information such as the intricate relationships among various targets, making it especially effective for multi-organ tasks. Conversely, the Transformer decoder, tailored to refine small and hard targets, is aptly suited for tumor segmentation. 
However, a combined approach of the Transformer encoder and decoder (Encoder+Decoder) does not offer further enhancements for either multi-organ or hepatic vessel segmentation. The reason may be attributed to their strengths may overlap or counteract, preventing synergistic benefits in these specific tasks. 
As a result, our subsequent experiments will predominantly employ the Encoder-only model for segmenting multiple objects and the Decoder-only model for segmentation tasks targeting smaller and more challenging tumors or lesions. 

\subsection{Deep Analysis for Transformer Decoder}

\subsubsection{Number of organ/tumor queries}
For successful training, the number of learnable queries must be at least equal to the number of classes ($i.e.,$ each class must have at least one query). However, when we varied the number of queries in the segmentation process, we observed that the performance of our 3D TransUNet with the Decoder-only configuration remains largely unaffected by this parameter. A detailed summary of these findings is presented in Table~\ref{tab:query-ablation}.

\subsubsection{Multi-scale CNN feature for updating queries}
A defining characteristic of our Transformer decoder is its integration of multi-scale features from the CNN decoder, which are rich in localization details. These features play a pivotal role in progressively refining the learnable queries through the synergy of cross-attention with localized multi-scale CNN representations. 
Our experimentation, as summarized in Table~\ref{tab:decoder-ablation}, encompasses the configuration of Decoder-only. The consistently observed performance enhancements, as compared to the baseline Transformer decoder—where the segmentation mask is computed by directly employing the dot product of the learned query and the last-layer CNN feature—underscore the indispensable nature of incorporating multi-scale CNN features in the query updating process.

\subsubsection{Coarse-to-fine refinement in Transformer decoder} 
\begin{figure*}[t]
    \centering
    \includegraphics[width=\textwidth]{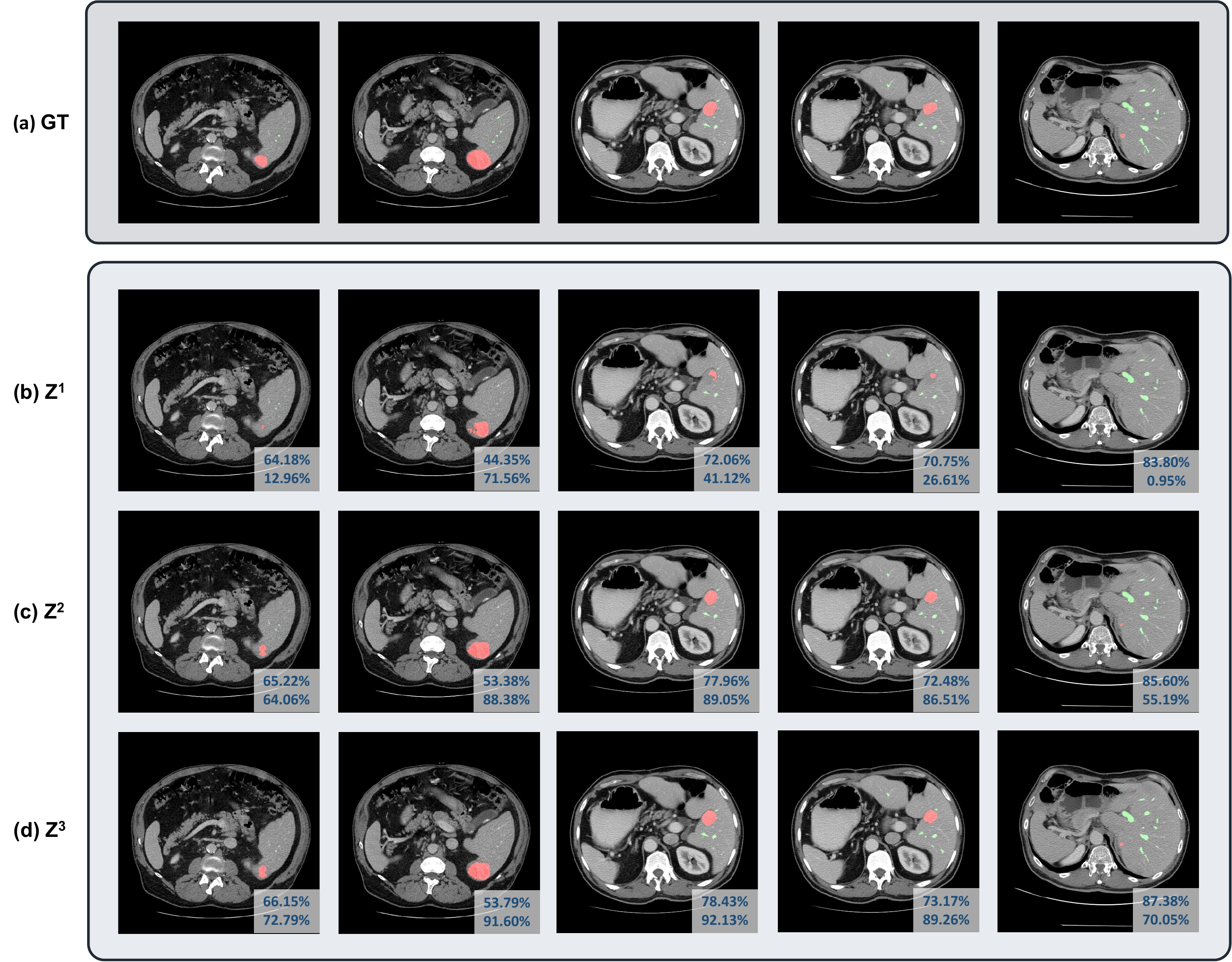}
    \caption{Visualizations of outputs from different iterations during coarse-to-fine refinement: (a) Groundtruth. (b-d) the segmentation mask at the first to the third iteration. Different columns represent different samples from MSD Vessel Dataset. The dice coefficients of vessels and tumors are indicated in the first and the second row of the lower right corner of each image respectively.}
    \label{fig:vis_c2f}
\end{figure*}

In each Transformer decoder layer, the coarse-to-fine refinement uses the predicted mask from the current iteration to constrain the cross-attention within the foreground region, therefore refining the organ queries at the next iteration. 
To demonstrate the effectiveness of this strategy, we have selected vessel tumor segmentation as a representative case study. This choice is motivated by the Transformer decoder's demonstrated proficiency in segmenting small targets, such as tumors or lesions. 
As illustrated in Table~\ref{tab:decoder-ablation}, 
the integration of coarse-to-fine refinement (masked cross-attention) consistently yielded enhanced results. 
For a more intuitive understanding, we provide a qualitative example in Fig.~\ref{fig:vis_c2f}, elucidating how this attention refines masks for intricate targets. From the first to the third iteration, the segmentation quality of the tumor has been significantly improved.

\begin{table*}
\centering
\footnotesize
\renewcommand\tabcolsep{1.15pt}
\caption{Comparison on the Synapse multi-organ CT dataset (average dice score \%, and dice score \% for each organ).
}
\resizebox{0.9\linewidth}{!}
{
\begin{tabular}{c|ccccccccccc}
\toprule[0.1em]
Scale & Method & Avg. Dice & Aorta & Gallbladder & Kidney (L) & Kidney (R) & Liver & Pancreas & Spleen & Stomach \\

\midrule[0.08em]
\multirow{6}{*}{2D (224$\times$224)}        & U-Net~\cite{ronneberger2015u}             & 74.68          & 84.18      & 62.84          & 79.19         & 71.29          & 93.35        & 48.23       & 84.41       & 73.92        \\
       & \scriptsize{Pyramid Attn} \cite{li2018pyramid} & 73.08      & 82.57 & 56.25	& 75.78	& 70.51	& 93.46	& 50.02	& 83.95	& 72.13        \\
        & \scriptsize{DeepLabv3+}~\cite{chen2018encoder} & 76.35   & 82.00	& 62.85	& 78.89	& 75.24	& 93.96	& 57.75	& 86.57	& 73.57        \\
       & UNet++ ~\cite{zhou2019unet++} & 76.65    & 86.93	& 63.69	& 77.86	& 68.29	& 93.91	& 59.23	& 87.81	& 75.49        \\
        & \scriptsize{AttnUNet}~\cite{schlemper2019attention} & 75.57        & 55.92      & 63.91          & 79.20         & 72.71          & 93.56       & 49.37        & 87.19       & 74.95        \\
& 
\textbf{TransUNet} & \textbf{77.48}          & 87.23        & 63.13          & 81.87          & 77.02          & 94.08      & 55.86         & 85.08      & 75.62        \\ 

\hline
\multirow{7}{*}{2D (512$\times$512)}        & U-Net~\cite{ronneberger2015u} & 81.34         & 89.69	& 69.98	& 83.08	& 74.13	& 95.10	& 67.73	& 90.50	& 80.51        \\
       & \scriptsize{Pyramid Attn}~\cite{li2018pyramid} & 80.08	 & 88.59	& 65.91	& 84.45	& 75.15	& 95.30	& 60.06	& 91.84	& 79.33       \\

       & \scriptsize{DeepLabv3+}~\cite{chen2018encoder} & 82.50	 & 88.79	& 72.16	& 88.13	& 79.52	& 95.58	& 65.97	& 90.02	& 79.87        \\
       & UNet++~\cite{zhou2019unet++} & 81.6  & 89.65	& 71.68	& 82.92	& 75.15 & 94.92	& 69.06	& 89.42	& 80.01        \\
     & AttnUNet~\cite{schlemper2019attention}   & 80.88    & 89.46 & 67.09 & 83.83 & 75.98 & 95.28 & 68.48 & 88.63 & 78.26  \\

& nnU-Net~\cite{isensee2021nnu}  & 82.92  & 91.55 & 73.43 & 82.74 & 73.61 & 96.01 & 71.81 & 94.29 & 79.94
\\
& \textbf{TransUNet} & \textbf{84.36} & 90.68 & 71.99       & 86.04      & 83.71      & 95.54 & 73.96    & 88.80  & 84.20   \\
\hline
\multirow{8}{*}{3D} & V-Net~\cite{milletari2016v}     & 68.81                & 75.34          & 51.87              & 77.10             & 80.75             & 87.84           & 40.05            & 80.56          & 56.98           \\
& DARR~\cite{fu2020domain}      & 69.77                & 74.74           & 53.77             & 72.31            & 73.24             & 94.08           & 54.18            & 89.90           & 45.96            \\

& nnU-Net~\cite{isensee2021nnu}      & 87.33       & 93.04        & 78.82      & 84.68          & 88.46         & 97.13        & 81.50       & 91.68        & 83.34             \\            
& CoTr~\cite{xie2021cotr} &85.72	&92.96	&71.09	&85.70	&85.71	&96.88	&81.28	&90.44	&81.74    \\
& nnFormer~\cite{zhou2023nnformer}  &85.32	&90.72	&71.67	&85.60	&87.02	&96.28	&82.28	&87.30	&81.69 \\
& VT-UNet~\cite{peiris2022robust} &70.72	&78.25	&44.76	&77.51	&78.16	&91.63	&45.18	&82.20	&68.04 \\
& Swin UNETR~\cite{hatamizadeh2021swin} &73.51	&82.94	&60.96	&80.41	&71.14	&91.55	&56.71	&77.46	&66.94  \\
& \textbf{TransUNet} & \textbf{88.11} & 92.97 & 81.15 & 85.76 & 87.47 & 97.03 & 81.76 & 93.39 & 85.31 \\
\bottomrule[0.1em]
\end{tabular}
}
\label{tab:synapse_comp_sota}
\end{table*}

\begin{table}[]
\small
\centering
\caption{Performance comparison on MSD vessel dataset with dice score metrics (\%). Experiments are conducted in five-fold cross-validation.}
\resizebox{0.85\linewidth}{!}
{
\begin{tabular}{c|ccc}
\toprule[0.1em]
Method       & Vessel & Tumor & Avg. Dice \\ 
\midrule[0.08em]
nnU-Net       & 63.71  & 68.36 & 66.04   \\
nnFormer~\cite{zhou2023nnformer}&63.21	&69.37	&66.29   \\
VT-UNet~\cite{peiris2022robust} &60.88	&59.82	&60.35 \\
Swin UNETR~\cite{hatamizadeh2021swin}  &57.65	&58.31	&57.98\\
\textbf{TransUNet} & 64.10  & 70.60 & \textbf{67.67}   \\ 
\bottomrule[0.1em]
\end{tabular}
}
\label{tab:vessel_comp_sota}
\end{table}

\subsection{Generalization to Other Tumor/Lesion Datasets}
We assess the generalizability of our 3D TransUNet across different imaging modalities and tasks, including brain metastases segmentation from MRI data (BraTS2023), pancreatic tumor (PDAC) and cyst segmentation. The outcomes are detailed in Table~\ref{tab:brats23_met} and Table~\ref{tab:pancreas_tumor}. Our findings confirm that the Encoder-only variant offers modest enhancements, whereas the Decoder-only configuration substantially boosts performance for tumor/lesion segmentation. For example, in brain metastatic lesion segmentation, the Encoder-only model provides a marginal enhancement, whereas the Decoder-only model contributes to a significant Dice improvement of $2.9\%$. This trend is even more pronounced for PDAC and cystic lesion segmentation. Specifically, Encoder-only model enhances the PDAC segmentation from 56.94\% to 58.38\%, while Decoder-only model pushes it further to 62.66\%. In the case of cystic lesions, Encoder-only model lifts nnU-Net's performance from 56.88\% to 57.98\%, with Decoder-only model further elevating it to 61.04\%. Consequently, the aggregate metrics of average Dice, Sensitivity, and Specificity all witness marked improvements with Decoder-only model.

\subsection{Comparison with State-of-the-arts}
We compare our TransUNet to previous 2D and 3D state-of-the-art methods on multi-organ segmentation and hepatic vessel tumor segmentation in Table~\ref{tab:synapse_comp_sota} and Table~\ref{tab:vessel_comp_sota}.
With the 2D version built on the U-Net architecture and the 3D variant grounded in the 3D nnU-Net framework, our TransUNet consistently outperforms other state-of-the-art methods, underscoring its efficacy across diverse U-Net frameworks.
As discussed above, leveraging the Transformer Encoder's ability to capture global organ relationships, we use the Encoder-only design for multi-organ segmentation. Conversely, given the Transformer Decoder's prowess in refining small targets, we opt for the Decoder-only setup for tumor segmentation.
Specifically, we compare TransUNet against a spectrum of methodologies, including: 1) 2D techniques such as U-Net~\cite{ronneberger2015u}, DeepLabv3+~\cite{chen2018encoder}, and UNet++~\cite{zhou2019unet++}, complemented by attention-augmented CNN methods like AttnUNet~\cite{schlemper2019attention} and Pyramid Attn~\cite{li2018pyramid}, evaluated across resolutions of \(224 \times 224\) and \(512 \times 512\); 2) 3D approaches like V-Net~\cite{milletari2016v}, DARR~\cite{fu2020domain}, and 3D nnU-Net~\cite{isensee2021nnu}, accompanied by cutting-edge Transformer-centric strategies including CoTR~\cite{xie2021cotr}, nnformer~\cite{zhou2023nnformer}, VT-UNet~\cite{peiris2022robust}, and Swin UNETR~\cite{hatamizadeh2021swin}.
As corroborated by the results in Table~\ref{tab:synapse_comp_sota}, TransUNet not only surpasses traditional CNN-based self-attention models but also outperforms numerous state-of-the-art Transformer-oriented techniques. For example, when benchmarked against recent state-of-the-art Transformer-based methods like CoTr and nnformer, our TransUNet achieves approximately a 10\% improvement in Dice scores for the challenging task of gallbladder segmentation and about a 3\% enhancement in overall segmentation

Notably, as evidenced in Table~\ref{tab:brats21}, our 3D TransUNet surpasses the top-ranked solution, nnU-Net-Large~\cite{luu2021extending}, from the BraTS2021 challenge, underscoring the robustness and efficacy of our proposed approach. 
\label{sec:comparison}

\input{table_figures/dsc_brats21}

\section{Conclusion}
While U-Net has been successful, its limitations in handling long-range dependencies have prompted the exploration of Transformer as an alternative architecture, exemplified by our previously developed TransUNet, harnessing the combined strengths of U-Net and Transformers. 
Our study extends the TransUNet architecture to a 3D network, building upon nnU-Net's foundation. Leveraging Transformers' capabilities in both encoder and decoder design, we introduce 1) A Transformer encoder that tokenizes CNN feature map patches, facilitating a richer extraction of global contexts; and
2) A Transformer decoder designed to adaptively refine segmentation regions, capitalizing on cross-attention mechanisms between candidate proposals and U-Net features.
Our investigations further illuminate that medical segmentation tasks have varying architectural preferences and offer insights into tailoring design configurations to suit these specific requirements.
Empirical results showcase our 3D TransUNet's superior performance in diverse medical segmentation tasks. Notably, our method excels in multi-organ, pancreatic tumor, hepatic vessel, and brain metastases segmentation. Additionally, we achieved notable results in the BraTS2021 challenge.

\vspace{-1ex}

\bibliographystyle{IEEEtran}
\bibliography{tmi}
\end{document}

%% file: table_figures/fig_decoder.tex
\begin{figure*}[t!]
    \centering
    \includegraphics[width=\textwidth]
    {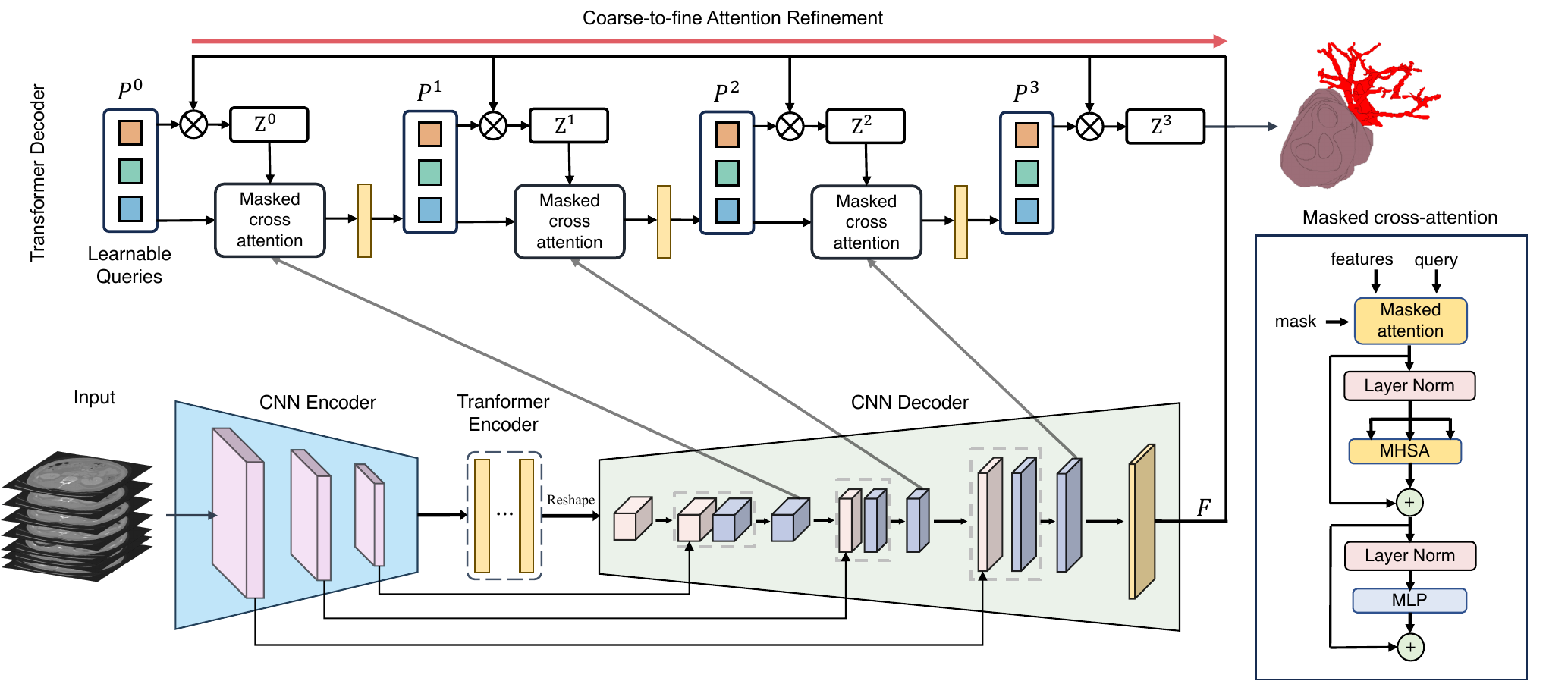}
    \caption{Overview of 3D TransUNet. We develop two opted architectural components: 1) The Transformer encoder where a CNN encoder is firstly used for local image feature extraction, followed by a pure Transformer encoder for global information interaction; and 2) the Transformer decoder that reframes per-pixel segmentation as mask classification using learnable queries, which are refined through cross-attention with CNN features, and employs a coarse-to-fine attention refinement approach for enhanced segmentation accuracy.}
    \label{fig:framework}
\end{figure*}

%% file: table_figures/tab_implement.tex
\begin{table}
\footnotesize
\renewcommand\tabcolsep{0.6pt}
\caption{implementation details including the architecture hyperparameter, training settings, and data augmentation.
}
\begin{tabular}{c|cccc}
\toprule[0.1em]
                    & Synapse                                       & MSD vessel                                      & BraTS                                    & Pancreas                                        \\ \midrule[0.08em]
category            & multi-organ                                   & organ\&tumor                                  & tumor                                        & organ\&tumor                                  \\
crop size           & 40$\times$224$\times$192                                    & 64$\times$192$\times$192                                      & 128$\times$128$\times$128                                  & 40$\times$224$\times$192                                      \\
batch size/gpu  & 2                                             & 2                                               & 2                                            & 2   \\
lr  & 8e-2                                             & 3e-4                                               & 3e-4                                            & 3e-4 \\
optimizer & sgd                                            & adamw                                               & adamw                                            & adamw \\
lr schedule &  cosine                        &  cosine                                           &  cosine                                 &  cosine 
\\
downsample   &          [4, 5, 5]                     &             [4, 5, 5]      &        {[}5, 5, 5{]}            & {[}3, 5, 5{]}                                   \\
num of query        & n/a                                           & 20                                              & 20                                           & 20                                              \\
C2F stage & 3                                             & 3                                               & 3                                            & 3                                               \\
augmentation        & \multicolumn{4}{l}{\begin{tabular}[c]{@{}l@{}}random rotation, scaling, flipping, white Gaussian noise, \\ Gaussian blurring, adjusting brightness and contrast, \\ simulation of low resolution, Gamma transformation\end{tabular}} \\ \bottomrule[0.1em]
\end{tabular}
\label{tab:implement}
\end{table}

%% file: table_figures/dsc_msdvessel.tex
\begin{table}[]
\tablestyle{3pt}{1.3}
\caption{Ablation of number of queries on MSD vessel dataset with dice score metrics (\%). Experiments are conducted in five-fold cross-validation.}
\resizebox{0.78\linewidth}{!}
{
\begin{tabular}{c|ccc}
\toprule[0.1em]
Number of Queries        & Vessel & Tumor & Avg. Dice \\ 
\midrule[0.08em]
5  & 64.75 & 70.32 & 67.53  \\
20 & 64.41 & 70.94 & 67.67   \\
40 & 64.32 & 70.41 & 67.37   \\ 
\bottomrule[0.1em]
\end{tabular}
}
\label{tab:query-ablation}
\end{table}

\begin{table}[]
\tablestyle{3pt}{1.3}
\caption{Ablation of transformer decoder on MSD vessel dataset with dice score metrics (\%). Experiments are conducted in five-fold cross-validation.}
\begin{tabular}{ccc|ccc}
\toprule[0.1em]
Transformer & \multirow{2}{*}{Multi-scale} & masked cross        & \multirow{2}{*}{Vessel} & \multirow{2}{*}{Tumor} & \multirow{2}{*}{Average} \\ 
decoder & &  attention\\
\midrule[0.08em]
\checkmark &\checkmark & \checkmark & 64.41 & 70.94 & 67.67 \\
\checkmark &\checkmark & & 64.37 & 70.71 & 67.54 \\
 \checkmark &  & & 64.19 & 69.89 & 67.04 \\
   &  & &63.71 & 68.36 & 66.04 \\
 \bottomrule[0.1em]
\end{tabular}
\label{tab:decoder-ablation}
\end{table}

%% file: table_figures/dsc_brats23_mets.tex
\begin{table}[]
\tablestyle{3pt}{1.3}
\caption{Performance comparison on BraTS2023 for Brain Metastasis Challenge with lesion-wise dice score metrics (\%). \\*masked cross attention is not used for the Decoder-only model for training stability. Experiments are conducted in five-fold cross-validation.}
\resizebox{0.75\linewidth}{!}
{
\begin{tabular}{c|cccc}
\toprule[0.1em]
Method       & ET & TC  & WT & Avg. Dice \\ 
\midrule[0.08em]
nnU-Net         &54.90  &58.67   &55.75 &56.44  \\
Encoder-only &54.79 &58.96 &56.05 &56.60 \\  
Decoder-only &56.80 &61.12  &60.09  &59.34  \\ 
\bottomrule[0.1em]
\end{tabular}
}
\label{tab:brats23_met}
\end{table}

%% file: table_figures/dsc_felix.tex
\begin{table}[]
\tablestyle{1.3pt}{1.3}
\caption{Performance comparison on in-house large-scale pancreatic tumor segmentation dataset.}
\begin{tabular}{c|cccc|cc}
\toprule[0.1em]
\multirow{2}{*}{Method}             & \multicolumn{4}{c|}{DSC}             & \multirow{2}{*}{Sensitivity} & \multirow{2}{*}{Specificity} \\ \cline{2-5}
      & Pancreas & PDAC  & Cyst  & Average &                              &                              \\ 
      \midrule[0.08em]

nnU-Net       & 83.8     & 56.94   & 56.88 & 65.97   & 89.34                    & 91.00                        \\
Encoder-only    & 83.77     & 58.38  & 57.98 & 66.71   & 91.71                 & 84.66                       \\
Decoder-only   & 85.35     & 62.66 & 61.04  & 69.69   & 89.94                   & 97.33                        \\ 
\bottomrule[0.1em]

\end{tabular}
\label{tab:pancreas_tumor}
\end{table}

%% file: table_figures/dsc_brats21.tex
\begin{table}[]
\tablestyle{3pt}{1.3}
\caption{Performance comparison on BraTS2021 challenge for brain tumor segmentation with dice score metrics (\%). Experiments are conducted in five-fold cross-validation.}
\resizebox{0.85\linewidth}{!}
{
\begin{tabular}{c|cccc}
\toprule[0.1em]
Method       & ET & TC  & WT & Avg. Dice \\ 
\midrule[0.08em]
nnU-Net       & 88.05  & 91.92  & 93.79   & 91.25  \\
AxialAttn~\cite{luu2021extending}   & 87.23  & 91.88  & 93.21    & 90.77  \\

nnUNet-Large~\cite{luu2021extending}  & 88.23  & 92.35 & 93.83  & 91.47  \\
\textbf{TransUNet} & 88.85  & 92.48  & 93.90   & \textbf{91.74}  \\ 
\bottomrule[0.1em]
\end{tabular}
}
\label{tab:brats21}
\end{table}

\vspace{-2ex}